\documentclass[journal]{IEEEtran}

\usepackage{amsmath}
\usepackage{amssymb}
\usepackage{xcolor}
\usepackage{graphicx}
\usepackage{booktabs}       

\usepackage{url}
\usepackage{subcaption}
\usepackage{subcaption}
\usepackage{float} 
\usepackage{multirow}     
\usepackage[table]{xcolor}  
\usepackage{colortbl}  
\usepackage[ruled,vlined,linesnumbered]{algorithm2e}
\usepackage{pgfplots}
\pgfplotsset{compat=1.17}

\definecolor{myorange}{RGB}{255, 240, 220} 
\definecolor{myblue}{RGB}{245, 252, 255} 
\definecolor{mygreen}{RGB}{245, 255, 245} 
\definecolor{mypink}{RGB}{255, 250, 252}   
\definecolor{myyellow}{RGB}{255, 254, 245}

\usepackage{caption}
\usepackage{cite}

\ifCLASSINFOpdf

\else

\fi

\begin{document}

\title{Dual-Stage Invariant Continual Learning under Extreme Visual Sparsity}

\author{Rangya~Zhang,
        Jiaping~Xiao,
        Lu~Bai,
        Yuhang~Zhang
        and~Mir~Feroskhan,~\IEEEmembership{Member,~IEEE}
\thanks{R. Zhang, J. Xiao, L. Bai, Y. Zhang and M. Feroskhan are with the School of Mechanical and Aerospace Engineering, 
Nanyang Technological University, Singapore 639798, Singapore 
(e-mail: rangya002@e.ntu.edu.sg; jiaping001@e.ntu.edu.sg; bailu@ntu.edu.sg; yuhang004@e.ntu.edu.sg; mir.feroskhan@ntu.edu.sg). 
\textit{(Corresponding author: Mir Feroskhan)}}}

\maketitle

\begin{abstract}
Continual learning seeks to maintain stable adaptation under non-stationary environments, yet this problem becomes particularly challenging in object detection, where most existing methods implicitly assume relatively balanced visual conditions. In extreme-sparsity regimes, such as those observed in space-based resident space object (RSO) detection scenarios, foreground signals are overwhelmingly dominated by background observations. Under such conditions, we analytically demonstrate that background-driven gradients destabilize the feature backbone during sequential domain shifts, causing progressive representation drift. This exposes a structural limitation of continual learning approaches relying solely on output-level distillation, as they fail to preserve intermediate representation stability. To address this, we propose a dual-stage invariant continual learning framework via joint distillation, enforcing structural and semantic consistency on both backbone representations and detection predictions, respectively, thereby suppressing error propagation at its source while maintaining adaptability. Furthermore, to regulate gradient statistics under severe imbalance, we introduce a sparsity-aware data conditioning strategy combining patch-based sampling and distribution-aware augmentation. Experiments on a high-resolution space-based RSO detection dataset show consistent improvement over established continual object detection methods, achieving an absolute gain of \textbf{+4.0 mAP} under sequential domain shifts.
\end{abstract}


\IEEEpeerreviewmaketitle

\section{Introduction}

\IEEEPARstart{C}{ontinual} learning (CL) in non-stationary environments requires maintaining stable representations under evolving visual conditions, particularly in complex visual tasks such as object detection. While most existing methods implicitly assume relatively balanced visual inputs, many real-world scenarios, such as aerial surveillance\cite{zhang2024npe} and long-range monitoring, exhibit extreme visual sparsity, where foreground signals occupy only a minute fraction of the visual field and are dominated by background observations. Such regimes are exemplified by space-based resident space object (RSO) detection, which provides a challenging setting for continual object detection under domain shifts and highly imbalanced visual conditions.

Current object detection systems~\cite{ren2015faster,Jocher_Ultralytics_YOLO_2023,carion2020end} achieve strong performance under stable and well-distributed training data, where targets occupy a reasonable portion of the visual field. However, under extreme foreground–background imbalance, this assumption breaks down. It induces a hierarchical sparsity pattern, where foreground scarcity persists both at the global image level and within localized candidate regions, severely limiting effective learning signals. Meanwhile, visual inputs continuously evolve over time with shifting viewpoints, leading to sequential domain shifts. Under these conditions, biased gradient accumulation arises, where weak target signals are dominated by background-driven updates. Consequently, the optimization process becomes skewed toward suppressing background variations rather than learning discriminative target features. This destabilizes intermediate feature representations, leading to accumulated representation drift across domains and gradual overwriting of previously learned features, ultimately triggering catastrophic forgetting~\cite{robins1995catastrophic}. Such structural instability underscores the need for Continual Object Detection (COD) frameworks that maintain stable representation evolution under extreme sparsity while adapting to sequential domain shifts.

\begin{figure*}[t]
    \centering
    \includegraphics[width=0.90\textwidth,height=0.85\textwidth]{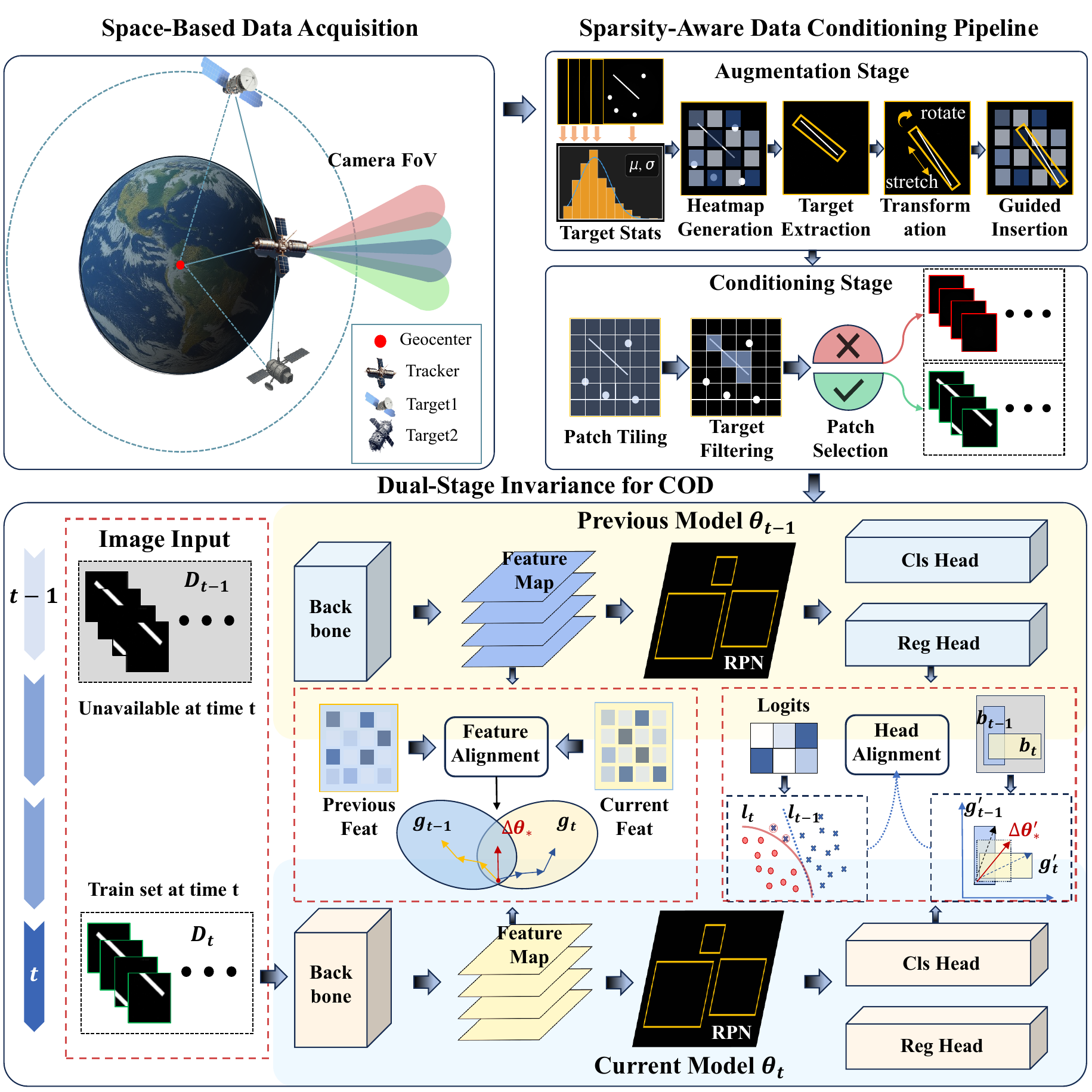}
    \caption{Overall framework of our method. The top left panel illustrates the space-based observation configuration. The top right shows our sparsity-aware data conditioning pipeline. The bottom panel depicts the dual-stage invariance in COD.}
    \label{fig:framework}
\end{figure*}

This challenge can be framed within the domain-incremental CL setting, where data distributions evolve while object categories remain fixed. Although stable adaptation in non-stationary and dynamic systems has been widely studied~\cite{zhang2022adaptive}, maintaining representation stability under such hierarchical sparsity conditions remains largely unexplored. Existing COD paradigms exhibit intrinsic limitations in this regime. Replay-based strategies~\cite{rebuffi2017icarl,wei2021incremental} and parameter-isolation methods~\cite{aljundi2017expert,wang2022dirichlet} preserve prior knowledge via data retention or model expansion, yet both implicitly depend on sufficiently informative samples to stabilize representation updates. Under background-dominated sparsity, effective target statistics are inherently scarce, rendering these mechanisms structurally fragile. Regularization and distillation-based approaches~\cite{kirkpatrick2017overcoming,zhang2020class} offer a more compact alternative by constraining parameter updates or output predictions, but still rely on the statistical stability of intermediate representations. As we analytically demonstrate, this assumption becomes unreliable under hierarchical sparsity: background-dominated gradients induce progressive representation drift from early layers, making head-level or parameter-level constraints insufficient to prevent forgetting. These observations motivate a principled framework that enforces dual-stage consistency over structural representations and detection semantics.

Motivated by the instability of representation evolution under extreme sparsity, we develop a COD framework tailored to domain-incremental adaptation under highly imbalanced visual streams. We design a dual-stage invariance mechanism that stabilizes representation dynamics across both intermediate backbone features and semantic detection predictions, thereby suppressing accumulated representation drift during sequential domain shifts. Furthermore, to ensure reliable learning under ultra-sparse and highly imbalanced visual inputs, we introduce a sparsity-aware data conditioning strategy that combines patch-based sampling with targeted distribution-aware augmentation to regulate gradient statistics and preserve effective learning signals. These components enable stable continual adaptation under extreme sparsity while maintaining strong detection performance on high-resolution space imagery. Our main contributions are summarized as follows:

\begin{enumerate}
  \item We reveal the mechanism of accumulated representation drift induced by extreme sparsity with hierarchical characteristics, providing insights into the breakdown of representation stability in COD under non-stationary and highly imbalanced visual streams.

  \item We propose a dual-stage invariance framework that jointly enforces structural consistency on backbone representations and semantic consistency on detection predictions, enabling stable representation evolution across sequential domain shifts.

  \item We develop a sparsity-aware data conditioning strategy based on patch sampling and distribution-aware augmentation to regulate optimization signals under severe foreground--background imbalance, supporting robust detection in ultra-high-resolution imagery, including space-based scenarios.
\end{enumerate}

The remainder of this paper is organized as follows. Section~\ref{sec:related} reviews continual learning and visual perception under non-stationary conditions. Section~\ref{sec:method} presents the proposed COD framework, including the sparsity-aware data conditioning pipeline and dual-stage invariance. Section~\ref{sec:dataset} analyzes the dataset, highlighting its extreme sparsity and distributional challenges. Section~\ref{sec:experiments} reports experimental results and ablation studies. Finally, Section~\ref{sec:conclusion} concludes the paper.

\section{Related Work}
\label{sec:related}
The growing demand for adaptive models in dynamic environments has brought CL to the forefront of machine learning research. Most recent works have explored various strategies to support CL under realistic constraints~\cite{lin2025multiple,lin2025machine}, leading to the development of regularization-based~\cite{li2021gopgan, zhang2020class, kirkpatrick2017overcoming,wen2024class}, replay-based~\cite{rebuffi2017icarl}, and distillation-based~\cite{li2017learning,aljundi2017expert} approaches. While differing in implementation, most methods share the common objective of mitigating catastrophic forgetting by constraining parameter updates or revisiting historical samples through memory replay. To date, the majority of CL methods have been evaluated in image classification settings, where task boundaries and label spaces are typically well defined, and optimization dynamics remain relatively stable.

Incremental image classification~\cite{rebuffi2017icarl} has driven the development of diverse CL approaches on benchmarks like CIFAR-100~\cite{krizhevsky2009learning} and ImageNet~\cite{deng2009imagenet}. Early methods such as LwF~\cite{li2017learning} utilized output distillation, while iCaRL~\cite{rebuffi2017icarl} combined exemplar replay with a nearest-mean classifier. Recent works explored simplified replay~\cite{lyu2025mitigating, yu2023contrastive,buzzega2020dark} or regularization strategies like EWC~\cite{kirkpatrick2017overcoming} to constrain parameter updates. While replay-based methods are effective, they incur memory overhead. Conversely, exemplar-free approaches~\cite{zhang2020class} avoid storage costs by modifying update directions or mimicking outputs, albeit with increased sensitivity to distributional shifts. Despite success in classification, extending these strategies to complex detection architectures with unstable optimization dynamics remains a significant challenge.

Driven by the demands of long-term model adaptation in real-world scenarios, CL has recently been extended to object detection tasks~\cite{shmelkov2017incremental, li2019rilod,liu2023continual}. However, the complex nature of object detection, involving both localization and classification with substantially larger model capacity, introduces unstable optimization dynamics, posing significant challenges to previously established CL methods. Recent efforts on COD have predominantly adopted knowledge distillation\cite{li2024esdb}, where selected components such as the detection head~\cite{shmelkov2017incremental}, intermediate features~\cite{li2019rilod}, or region proposal networks (RPN)~\cite{hao2019end} are constrained to preserve prior knowledge. While effective in relatively structured settings, these single-stage distillation schemes are insufficient to suppress the progressive representation drift induced by severe sparsity and domain shifts. Alternatively, exemplar-based replay strategies~\cite{liu2023continual} have also been explored, yet often suffer from overfitting to limited replay samples and struggle to scale under bandwidth or memory constraints. Motivated by these limitations, we develop a dual-stage invariance strategy that explicitly constrains both backbone representations and detection predictions, enabling robust knowledge retention without relying on replay buffers or access to previous data, and supporting deployment in memory-constrained scenarios.

Categorized by task type, CL is divided into class-incremental, task-incremental, and domain-incremental settings~\cite{van2022three}. While the former two involve evolving label spaces or require task identifiers, domain-incremental learning retains a fixed label space ($\mathbb{P}(Y_t) = \mathbb{P}(Y_{t+1})$) but confronts significant input distributional shifts ($\mathbb{P}(X_t) \neq \mathbb{P}(X_{t+1})$) without access to task identity. Most existing COD methods focus on the class-incremental paradigm; however, many real-world visual scenarios are inherently domain-incremental, driven by variations in viewpoints, illumination, and noise rather than new categories, and are often further complicated by extreme target sparsity and resource constraints. Consequently, our work targets this underexplored setting, aiming for exemplar-free adaptation that explicitly addresses these challenges.

Despite the increasing attention to COD and the need for robust domain adaptation, most state-of-the-art methods are still benchmarked on datasets like PASCAL VOC~\cite{everingham2010pascal} and MS COCO~\cite{lin2014microsoft}, where task differences are manually created by introducing new classes rather than arising from natural domain shifts. However, in unconstrained physical environments\cite{xiao2024toward} (e.g., open-world perception), visual streams often exhibit extreme sparsity, where targets occupy only a minute fraction of the visual field and are dominated by background observations. Such conditions introduce significant challenges, including severe foreground--background imbalance, limited effective learning signals, and strong variations in viewpoint and noise, leading to complex and unstable domain shifts. These characteristics fundamentally challenge the stability of representation learning and continual adaptation.

Overall, existing COD approaches remain inadequate for addressing practical domain-incremental deployments, particularly under severe sparsity and complex domain shifts. Such conditions are prevalent across a range of visual perception tasks, including aerial imagery, long-range surveillance, and satellite-based observation, where foreground signals are limited, and distributions evolve over time. These challenges necessitate a domain-incremental detection framework that can robustly adapt to evolving visual conditions without relying on exemplar storage or access to previous data. In particular, space-based imagery represents a highly challenging instance of this problem, characterized by extreme sparsity and continuously changing observation conditions.

\begin{figure*}[t]
    \centering
    \includegraphics[width=0.90\linewidth]{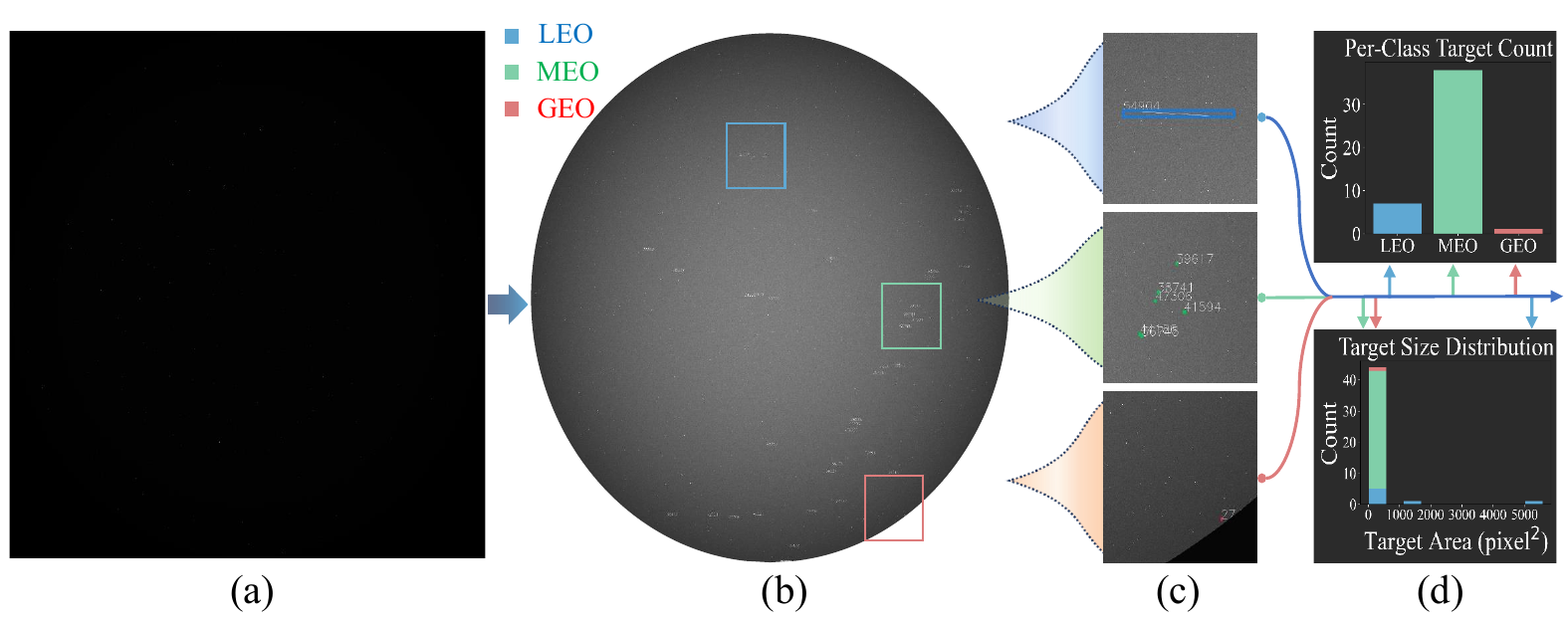}
    \caption{Overview of dataset characteristics. 
    (a) Raw $4418\times4418$ 16-bit grayscale image with sparse targets barely visible. 
    (b) Reference image after contrast enhancement for visualization. 
    (c) Zoomed regions highlighting LEO (blue), MEO (green), and GEO (red) RSOs. 
    (d) Per-class target count (top) and target size distribution (bottom) for the shown example.}
    \label{fig:dataset_overview}
\end{figure*}

\section{Methodology}
\label{sec:method}
Unlike conventional object detection problems, where foreground targets are abundant and data distributions are relatively stable, advanced visual perception systems operating in unconstrained environments (e.g., space-based observation) are significantly challenged by extreme target sparsity and domain shifts caused by environmental complexity~\cite{zhang2024astro}. To address these issues, this section introduces a COD framework specialized for extreme visual sparsity, aiming to improve both accuracy and generalizability across non-stationary visual streams.

We begin by formally defining the domain incremental learning problem in Sec.~\ref{sec:problem_formulation}. Data processing and augmentation strategies are described in Sec.~\ref{sec:data_processing}, and Sec.~\ref{sec:framework} concludes with the presentation of our COD framework.

\subsection{Problem Formulation}
\label{sec:problem_formulation}

CL aims to incrementally adapt models to new data distributions while retaining knowledge from previous ones~\cite{parisi2019continual}. This work considers domain incremental learning, a CL scenario where datasets from distinct domains are encountered sequentially, each exhibiting significant distributional shifts.

Let $\mathcal{D} = \{\mathcal{D}^{(1)}, \mathcal{D}^{(2)}, \dots, \mathcal{D}^{(T)}\}$ denote a set of domains, where each domain $\mathcal{D}^{(t)}$ is associated with a joint data-label distribution $P^{(t)}(X, Y)$. We assume domain-wise disjointness, i.e., $\mathcal{D}^{(i)} \cap \mathcal{D}^{(j)} = \emptyset$ for $i \neq j$. At each time step $t \in \{1, \dots, T\}$, only samples $(X^{(t)}_i, Y^{(t)}_i) \sim P^{(t)}(X, Y)$ from the current domain $\mathcal{D}^{(t)}$ are accessible for training, and data from other domains are unavailable.

For the object detection setting, let $I_i^{(t)} \in X^{(t)}$ be an image sampled from domain $\mathcal{D}^{(t)}$, and $Y^{(t)}$ be the corresponding set of bounding box annotations. We aim to train an object detector $f_\theta$ parameterized by $\theta$, that performs well across all seen domains despite only having access to one domain at a time.
A key challenge arises due to covariate shift across domains: while the label space remains constant ($P^{(t)}(Y) = P^{(t+1)}(Y)$), the input distributions differ significantly ($P^{(t)}(X) \neq P^{(t+1)}(X)$). Under this constraint, our learning objective is to minimize the expected detection loss across all observed domains:
\begin{equation}
\min_\theta \quad \frac{1}{T} \sum_{t=1}^{T} \mathbb{E}_{(X, Y) \sim \mathcal{D}^{(t)}} \left[ \mathcal{L} \left(f_\theta(X), Y \right) \right]
\end{equation}
where $\mathcal{L}$ is the task-specific detection loss (e.g., classification + regression), and $T$ is the number of domains seen so far.

This formulation captures the core challenge of COD under extreme visual sparsity: achieving high adaptability to new visual domains while preserving detection capabilities in previously learned domains, without accessing historical data.

\subsection{Sparsity-Aware Data Conditioning Strategy}
\label{sec:data_processing}
To address the extreme sparsity and inter-domain imbalance inherent to continual learning in extremely sparse visual environments, we propose a sparsity-aware data conditioning strategy comprising an augmentation module $\mathcal{A}(\cdot)$ and a patch sampling module $\mathcal{P}(\cdot)$, as illustrated in Fig.~\ref{fig:framework}, ultimately yielding the final training set $\mathcal{D}^*$. 

\begin{equation}
\mathcal{D}^* = \mathcal{P}(\mathcal{A}(X))
\end{equation}

In this formulation, $\mathcal{A}$ addresses inter-domain target imbalance for balanced learning, while $\mathcal{P}$ filters informative regions to improve efficiency and scalability.

\subsubsection{Distribution Rebalancing via Augmentation}
Standard CL assumes balanced target counts across domains~\cite{parisi2019continual}, whereas space imagery often exhibits large disparities that lead to biased learning. To address this, we propose an augmentation pipeline $\mathcal{A}(\cdot)$ that equalizes target distributions by injecting transformed patches.

For domain $\mathcal{D}^{(t)}$, we model the target count distribution $\mathcal{C}^{(t)}$ as a Gaussian and capture the spatial distribution of target centers via a normalized heatmap $\mathcal{S}^{(t)}$:
\begin{equation}
\mathcal{C}^{(t)} \sim \mathcal{N}(\mu^{(t)}, {\sigma^{(t)}}^2), \quad \text{with } \mu^{(t)} = \frac{1}{|X^{(t)}|} \sum_{i=1}^{|X^{(t)}|} c_i
\end{equation}
\begin{equation}
\mathcal{S}^{(t)}(x, y) = \frac{1}{Z} \sum_{i=1}^{|X^{(t)}|} \sum_{j=1}^{c_i} \exp\left( -\frac{\| (x, y) - \mathbf{p}_{ij}^{(t)} \|^2}{2\sigma_s^2} \right)
\end{equation}
where $c_i$ is the target count in image $I_i^{(t)}$, $\mathbf{p}_{ij}^{(t)}$ denotes target centers, and $\sigma_s$ controls spatial smoothing.

Target patches $\mathcal{B}^{(t)} = \{b_k^{(t)}\}$ are extracted and subjected to affine transformations $\mathcal{T}$ (rotation $\theta$, scaling $\lambda$) to preserve semantic consistency:
\begin{equation}
b_k^{\prime(t)} = \mathcal{T}(b_k^{(t)}), \quad \text{with } \mathcal{T} \in \text{Affine}(\theta, \lambda)
\end{equation}
Finally, we sample insertion locations $(x^*, y^*) \sim \mathcal{S}^{(t)}$ and blend the patch using a max-pixel strategy:
\begin{equation}
I_i^{(t)}(x, y) = \max\left( I_i^{(t)}(x, y), b_k^{\prime(t)}(x - x^*, y - y^*) \right)
\end{equation}
This process repeats until the expected target count $\mathbb{E}[\mathcal{C}^{(t)}]$ approximates a predefined constant $K$.

\subsubsection{Signal Conditioning via Sampling}
CNN-based detectors~\cite{ren2015faster,Jocher_Ultralytics_YOLO_2023,lin2017focal} $F_{\text{OD}}(\mathbf{x}; \theta)$ are typically trained on full-resolution images $I_i^{(t)} \in X^{(t)}$. While this enables global context modeling, it also results in overwhelming background content and heavy memory usage. To mitigate these challenges, we introduce a conditioning pipeline $\mathcal{P}(\cdot)$ adapted to sparse space imagery. Each image $I_i^{(t)} \in X^{(t)}$ is divided into $N_i^{(t)}$ patches $\mathcal{D}_i^{(t)} = \{x_{i}^{(t, j)}\}_{j=1}^{N_i^{(t)}}$, where $x_{i}^{(t, j)} \in \mathbb{R}^{H' \times W'}$. Non-informative patches with no target content are discarded by a filter $\phi$, yielding the final training set.
\begin{equation}
\begin{aligned}
\mathcal{D}^* &= \bigcup_{t=1}^{T} \bigcup_{i=1}^{|X^{(t)}|} \left\{x_{i}^{(t, j)} \mid \phi(x_{i}^{(t, j)}) = 1\right\}, \\
\mathcal{D}^* \cup \tilde{\mathcal{D}} &= \bigcup_{t=1}^{T} \bigcup_{i=1}^{|X^{(t)}|} \mathcal{D}_i^{(t)}
\end{aligned}
\end{equation}

This patch-based filtering not only reduces redundant background regions but also improves training efficiency and alleviates bandwidth constraints, making it well-suited for resource-limited edge visual perception systems.

\begin{algorithm}[h]
\caption{Sparsity-Aware Data Conditioning}
\label{alg:dataprep}
\KwIn{Raw dataset $\mathcal{D} = \{X^{(1)}, \dots, X^{(T)}\}$}
\KwOut{Processed training set $\mathcal{D}^*$}
\BlankLine

\ForEach{domain $X^{(t)} \in \mathcal{D}$}{
  Estimate target count distribution $\mathcal{C}^{(t)} \sim \mathcal{N}(\mu^{(t)}, \sigma^{2(t)})$\;
  Generate spatial heatmap $\mathcal{S}^{(t)}(x,y)$ from annotated target centers\;

  Extract target patches $\mathcal{B}^{(t)} = \{b_k^{(t)}\}_{k=1}^{N^{(t)}}$ from $X^{(t)}$\;
  \ForEach{patch $b_k^{(t)} \in \mathcal{B}^{(t)}$}{
    Apply light transformation: $b_k' = \mathcal{T}(b_k^{(t)})$\;
    Sample insertion location: $(x^*, y^*) \sim \mathcal{S}^{(t)}$\;
    Blend into image: $I(x, y) \leftarrow \max\big(I(x, y), b_k'(x - x^*, y - y^*)\big)$\;
  }
}

\ForEach{image $I_i^{(t)} \in X^{(t)}$}{
  Divide image into patches: $\mathcal{D}_i^{(t)} = \{x_i^{(t,j)}\}_{j=1}^{N_i^{(t)}}$\;
  \ForEach{patch $x_i^{(t,j)} \in \mathcal{D}_i^{(t)}$}{
    \If{$\phi(x_i^{(t,j)}) = 1$}{
      Add to final training set: $\mathcal{D}^* \leftarrow \mathcal{D}^* \cup \{x_i^{(t,j)}\}$\;
    }
  }
}
\Return{$\mathcal{D}^*$}
\end{algorithm}

\subsection{Dual-Stage Invariance for COD}
\label{sec:framework}

To address domain-incremental RSO detection, we propose a dual-stage invariance framework enforcing multi-level preservation across structural and semantic representations. As shown in Fig.~\ref{fig:framework}, our method builds upon the standard Faster R-CNN~\cite{ren2015faster}, augmented with a patch-based conditioning module and a dual-stage invariance strategy to stabilize learning under cross-domain shifts.

At step $t$, the detector is initialized from the previous model $\theta_{t-1}$, which also serves as a frozen teacher. Inputs from $\mathcal{D}_t$ are processed by both models to yield backbone features and RoI outputs. To preserve knowledge under severe sparsity, we introduce two constraints. \emph{Feature alignment} regularizes backbone representations $\phi_{\text{bb}}(x;\theta_t)$ against $\phi_{\text{bb}}(x;\theta_{t-1})$, suppressing drift and background-dominated gradient fluctuations. \emph{Head alignment} matches RoI-level predictions $\psi(x,r;\theta_t)$ with $\psi(x,r;\theta_{t-1})$, ensuring stable classification and regression. By jointly constraining backbone and outputs, this mechanism reshapes the optimization geometry and mitigates accumulated deviation from low-level representations to high-level predictions. A theoretical justification from a hierarchical sparsity perspective is provided in Appendix A-A and Appendix A-B.

This section presents the core motivation behind our framework (Sec.~\ref{sec:motivation}), and introduces the proposed invariance constraints for distillation (Sec.~\ref{sec:constraint}) that enable continual adaptation without catastrophic forgetting.

\subsubsection{Motivation and Challenges}
\label{sec:motivation}
Unlike class-incremental learning, which expands the label space $\mathcal{Y}$, domain-incremental detection under extreme sparsity retains a fixed $\mathcal{Y}$ but confronts significant input distribution shifts $\mathcal{D}_t$ caused by varying observation angles and illumination. These shifts induce substantial representation drift, which is further exacerbated by extreme target sparsity and background dominance in unconstrained visual streams.

\noindent
(i) \textit{Feature Representation Shift}: Overwhelming background gradients drive backbone parameters $\theta_{\text{bb}}$ to diverge from the source domain manifold, causing the feature extractor to lose discriminability for previously learned objects.

\noindent
(ii) \textit{Amplified Error Propagation}: Detection heads are highly sensitive to input feature perturbations. Minor shifts in backbone features are non-linearly amplified through the RPN and RoI heads, causing severe instability in localization and classification.

\noindent
(iii) \textit{Optimization Conflict}: Standard head-only distillation constrains outputs while allowing input features to shift freely. This misalignment between drifting inputs and fixed output targets creates an ill-posed optimization problem, impeding convergence.

We denote the shift-induced representation difference at stage $l$ as:
\begin{equation}
\Delta \mathbf{f}_l^{(t)} = \phi_l(\mathbf{x}; \theta_t) - \phi_l(\mathbf{x}; \theta_{t-1}) = \delta_l^{(t)}, \quad l \in \{\text{bb}, \text{RoI}\}
\end{equation}
where $\delta_l^{(t)}$ captures the cross-domain deviation. Unconstrained optimization on $\mathcal{D}_t$ amplifies such deviations, particularly under severe sparsity where the effective gradient is dominated by background components. This leads to progressive accumulation of representation drift, deteriorating semantic consistency and causing catastrophic forgetting.

Furthermore, space-based detection exhibits a pronounced hierarchical sparsity property: targets occupy a tiny fraction of pixels even within positive proposals. Consequently, optimization signals are overwhelmed by background statistics, yielding a low signal-to-noise regime where naive fine-tuning suffers from unstable gradient directions. Appendix A-B provides a formal analysis of this pathology.

\subsubsection{Limitations of Head-only Distillation}
\label{sec:limitation}

Most COD methods adopt head-only distillation to align outputs between the student $\theta_t$ and frozen teacher $\theta_{t-1}$:
\begin{equation}
\begin{split}
\mathcal{L}_{\text{RoI}} = \sum_{r \in \mathcal{R}} \bigg[ & \mathrm{KL}(p_{\theta_{t-1}}^{(r)} \| p_{\theta_t}^{(r)}) \\
& + \beta\, \mathrm{SmoothL1}(b_{\theta_t}^{(r)} - b_{\theta_{t-1}}^{(r)}) \bigg]
\end{split}
\label{eq:roi_distill}
\end{equation}
This formulation implicitly assumes proposal invariance ($\mathcal{R}_{\theta_t} \approx \mathcal{R}_{\theta_{t-1}}$). However, in domain-incremental RSO detection, shifts induce backbone drift and geometric misalignment (Sec.~\ref{sec:motivation}), causing structural discrepancies in proposal sets. This yields mismatched gradient statistics, rendering RoI supervision noisy under severe sparsity.

Critically, head-only distillation fails to regularize upstream backbone drift. Let $\mathbf{y} = h_{\boldsymbol{\psi}}( f_{\boldsymbol{\phi}}(\mathbf{x}) )$. Under hierarchical sparsity, backbone drift dominates ($\|\Delta f_{\boldsymbol{\phi}}\| \gg 0$). Even with strong head regularization ($\|\Delta h_{\boldsymbol{\psi}}\| \approx 0$), the output deviation remains unbounded:
\begin{equation}
\|\Delta \mathbf{y}\| \approx \left\| \nabla_{f} h_{\boldsymbol{\psi}} \right\| \cdot \|\Delta f_{\boldsymbol{\phi}}\| \gg \varepsilon
\label{eq:cascade_bound}
\end{equation}
This accumulated deviation (proven in Appendix A-D) confirms that output-level regularization alone is structurally insufficient, motivating our dual-stage strategy.

\subsubsection{Dual-Stage Invariance}
\label{sec:constraint}

Motivated by the analysis in Secs.~\ref{sec:motivation} and~\ref{sec:limitation}, we propose a dual-stage distillation strategy that jointly constrains feature-level and semantic-level deviations, thereby suppressing accumulated error propagation under extreme sparsity and domain shift. The proposed framework imposes complementary invariance objectives at the backbone and RoI head, enabling robust replay-free continual adaptation.

\paragraph{Structural Invariance: Stabilizing Backbone Representations.}
To stabilize low-level representations and mitigate background-induced perturbations, we introduce a structural invariance objective that aligns intermediate backbone features between the current model $\theta_t$ and the frozen teacher $\theta_{t-1}$. Specifically, for a given input $\mathbf{x}$, we encourage consistency of multi-scale backbone features:
\begin{equation}
\mathcal{L}_{\text{feat}} 
= \sum_{l \in \mathcal{L}_{\text{bb}}} 
\left\| 
\phi_l(\mathbf{x}; \theta_t) - \phi_l(\mathbf{x}; \theta_{t-1}) 
\right\|_2^2
\label{eq:feat_distill}
\end{equation}
where $\phi_l(\cdot;\theta)$ denotes the feature representation at backbone stage $l$, and $\mathcal{L}_{\text{bb}}$ indexes the selected layers.

This constraint regularizes intermediate backbone features, suppressing cross-domain drift accumulated across layers. Under extreme sparsity (Sec.~\ref{sec:motivation}), feature consistency attenuates background-dominated perturbations and stabilizes the optimization trajectory. A rigorous theoretical interpretation of this stabilization effect, including its spectral view and gradient-SNR implications, is provided in Appendix A-D.

\paragraph{Semantic Invariance: Stabilizing Detection Predictions.}
While feature-level regularization stabilizes intermediate representations, preserving high-level semantic knowledge requires output-level supervision at the detection head. We therefore adopt the RoI-level distillation loss in Eq.~\eqref{eq:roi_distill} to align both classification outputs and bounding-box regressions between $\theta_t$ and $\theta_{t-1}$. Unlike conventional head-only distillation (Sec.~\ref{sec:limitation}), this semantic supervision is applied on backbone features already stabilized by Eq.~\eqref{eq:feat_distill}, which improves the reliability of teacher guidance under domain shift and reduces gradient misalignment.

\begin{algorithm}[h]
\caption{Continual Adaptation with Dual-Stage Invariance (Step $t$)}
\label{alg:training}
\KwIn{Student $\theta_t$ initialized from $\theta_{t-1}$, frozen teacher $\theta_{t-1}$, mini-batch $\mathcal{B}_t=\{(\mathbf{x}_i,\mathbf{y}_i)\}$, weights $\lambda_f,\lambda_h$, learning rate $\eta$}
\KwOut{Updated student parameters $\theta_t$}

\BlankLine
\tcc{0. Freeze teacher}
Freeze $\theta_{t-1}$ and detach all teacher outputs (stop-gradient)\;

\BlankLine
\tcc{1. Forward pass (teacher \& student)}
Compute teacher backbone features $\{\phi_l(\mathbf{x}_i;\theta_{t-1})\}$\;
Compute student backbone features $\{\phi_l(\mathbf{x}_i;\theta_t)\}$ and generate RoIs $\mathcal{R}$ using the student RPN\;
Evaluate teacher RoI outputs $\{p_{\theta_{t-1}}^{(r)}, b_{\theta_{t-1}}^{(r)}\}_{r\in\mathcal{R}}$ and student RoI outputs $\{p_{\theta_t}^{(r)}, b_{\theta_t}^{(r)}\}_{r\in\mathcal{R}}$\;

\BlankLine
\tcc{2. Detection loss on current domain}
$\mathcal{L}_{\text{det}}^{(t)} \leftarrow \mathcal{L}_{\text{det}}(\theta_t;\mathcal{B}_t)$\;

\BlankLine
\tcc{3. Dual-stage Invariance losses}
$\mathcal{L}_{\text{feat}} \leftarrow \sum\limits_{l \in \mathcal{L}_{\text{bb}}}\left\| \phi_l(\mathbf{x};\theta_t) - \phi_l(\mathbf{x};\theta_{t-1}) \right\|_2^2$\;
$\mathcal{L}_{\text{RoI}} \leftarrow \sum\limits_{r \in \mathcal{R}}\Big[\mathrm{KL}(p_{\theta_{t-1}}^{(r)}\|p_{\theta_t}^{(r)}) + \beta\,\mathrm{SmoothL1}(b_{\theta_t}^{(r)}-b_{\theta_{t-1}}^{(r)})\Big]$; 

\BlankLine
\tcc{4. Unified objective}
$\mathcal{L}_{\text{total}}^{(t)} \leftarrow \mathcal{L}_{\text{det}}^{(t)} + \lambda_f\,\mathcal{L}_{\text{feat}} + \lambda_h\,\mathcal{L}_{\text{RoI}}$\;

\BlankLine
\tcc{5. Parameter update (SGD/Adam)}
$\theta_t \leftarrow \theta_t - \eta \nabla_{\theta_t}\mathcal{L}_{\text{total}}^{(t)}$\;

\BlankLine
\Return $\theta_t$\;
\end{algorithm}

\begin{figure*}[htbp]
    \centering
    \includegraphics[width=0.90\linewidth]{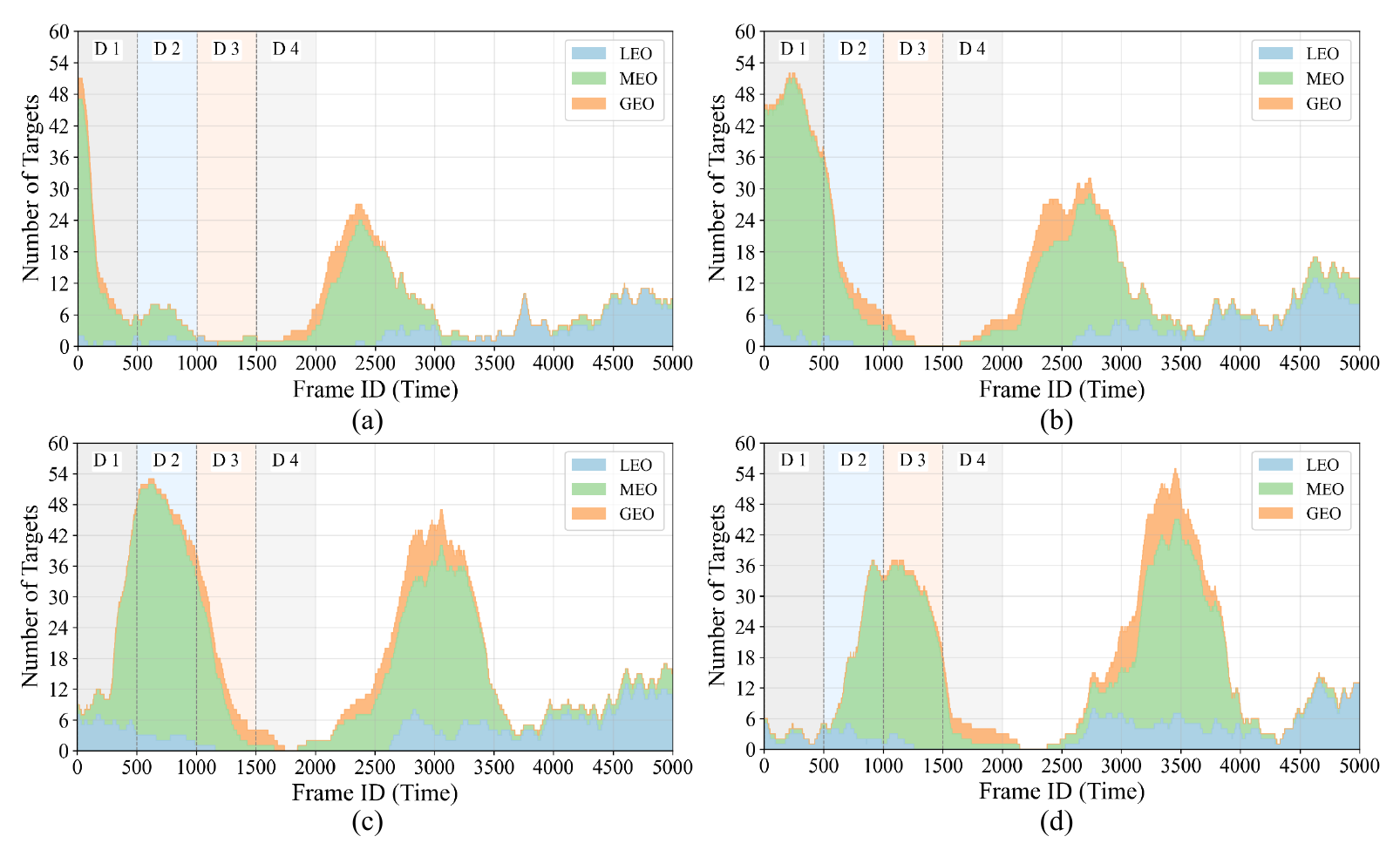}
    \caption{Temporal target distributions across all four camera datasets. Each subplot shows the per-frame target count over time, with LEO/MEO/GEO class breakdown. Background regions labeled D 1 to D 4 indicate domain divisions.}
    \label{fig:target_density_curve}
\end{figure*}

\paragraph{Necessity of dual-stage invariance.}
Feature and head distillation play complementary roles in continual RSO detection. As shown in Sec.~\ref{sec:limitation}, head-only regularization cannot prevent accumulated deviation when backbone deviation $\|\Delta f_{\boldsymbol{\phi}}\|$ is large (cf. Eq.~\eqref{eq:cascade_bound}). In contrast, dual-stage invariance directly suppresses the root cause of this accumulated effect by jointly constraining backbone and head deviations. Specifically, feature distillation reduces representation drift via Eq.~\eqref{eq:feat_distill}, which in turn yields a bounded prediction deviation:
\begin{equation}
\|\Delta \mathbf{y}\|_{\text{dual-stage}}
\;\le\;
\left\| \nabla_f h_{\boldsymbol{\psi}} \right\| \cdot \epsilon_{\text{feat}}
\;+\;
\epsilon_{\text{head}}
\label{eq:dual_bound_main}
\end{equation}
where $\epsilon_{\text{feat}}$ and $\epsilon_{\text{head}}$ denote the residual deviations after feature-level and head-level regularization, respectively. Although these residuals do not vanish, dual-stage invariance substantially reduces both terms compared to head-only supervision, i.e., $\epsilon_{\text{feat}} \ll \|\Delta f_{\boldsymbol{\phi}}\|$, thereby yielding significantly smaller output drift. A detailed derivation of this irreplaceability result under hierarchical sparsity is provided in Appendix A-D.

\subsubsection{Unified Objective and Optimization}
\label{sec:objective}

To jointly enforce structural and semantic invariance while enabling stable adaptation to the current domain $\mathcal{D}_t$, we formulate a unified optimization framework that integrates detection learning and dual-stage knowledge preservation. Rather than treating distillation as auxiliary supervision, we interpret it as a mechanism to reshape the optimization geometry and regulate gradient flow under severe domain shift and hierarchical sparsity.

\paragraph{Preservation functional.}
Following the analysis in Secs.~\ref{sec:motivation}–\ref{sec:constraint}, we define a multi-level preservation functional that penalizes deviations from the previous model $\theta_{t-1}$ across both backbone representations and RoI-level predictions:
\begin{equation}
\mathcal{P}(\theta_t;\theta_{t-1})
= 
\lambda_f \, \mathcal{L}_{\text{feat}}
+ 
\lambda_h \, \mathcal{L}_{\text{head}}
\label{eq:preserve_func}
\end{equation}
where $\mathcal{L}_{\text{feat}}$ and $\mathcal{L}_{\text{head}}$ are defined in Eqs.~\eqref{eq:feat_distill} and~\eqref{eq:roi_distill}, respectively. 
This functional imposes structured regularization over critical representation subspaces, discouraging parameter updates that induce excessive cross-domain deviation while preserving sufficient flexibility for learning new domain characteristics.

\begin{figure*}[htbp]
    \centering
    \includegraphics[width=0.90\linewidth]{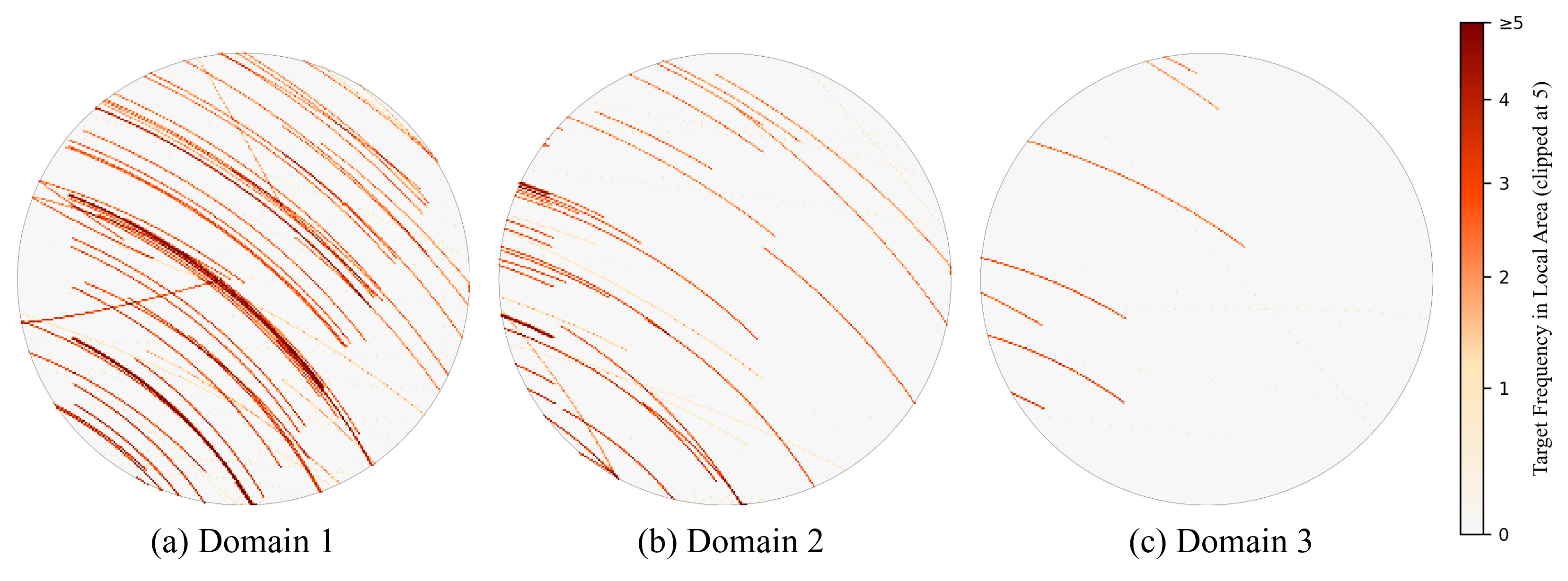}
    \caption{Spatial distributions of target centers across three consecutive subsets of 500 images each (Domain 1 to Domain 3). Each heatmap reflects the local density of targets per grid cell, clipped at a maximum count of 5.}
    \label{fig:target_heatmap}
\end{figure*}

\begin{figure*}[t]
    \centering
    \begin{subfigure}[b]{0.32\textwidth}
        \centering
        \includegraphics[width=\linewidth]{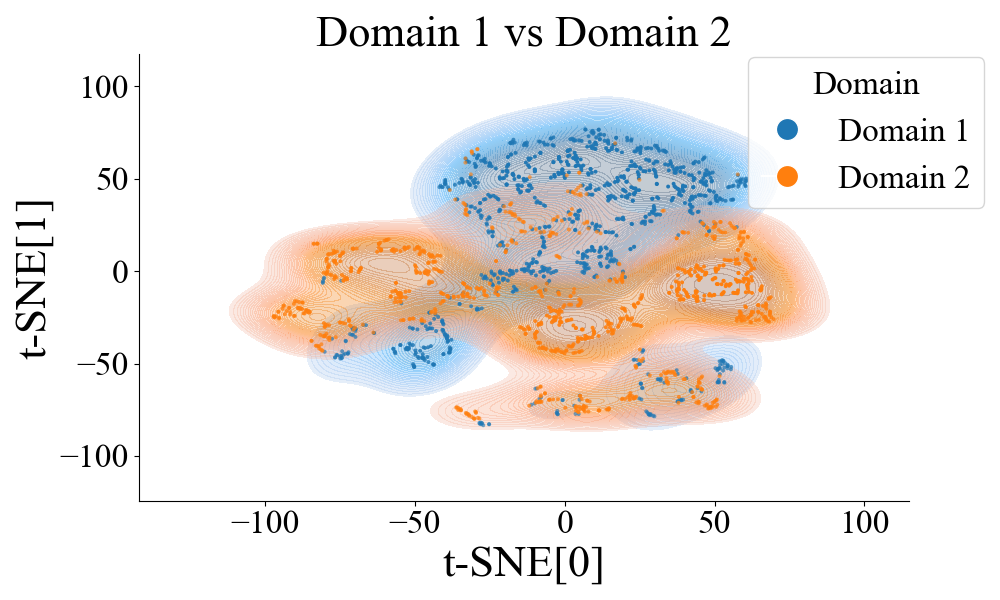}
        \caption{Domain 1 vs 2}
    \end{subfigure}
    \hfill
    \begin{subfigure}[b]{0.32\textwidth}
        \centering
        \includegraphics[width=\linewidth]{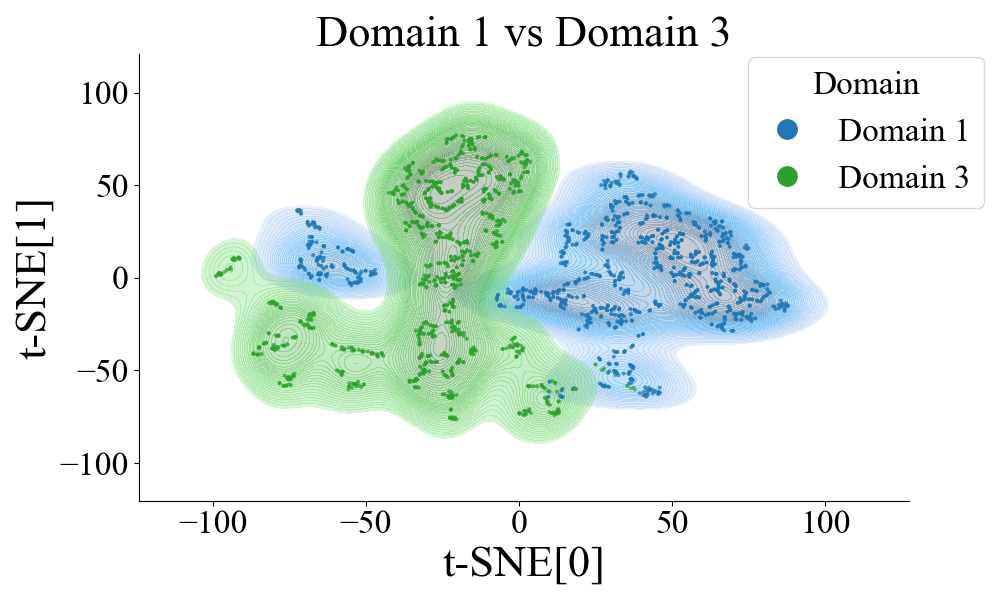}
        \caption{Domain 1 vs 3}
    \end{subfigure}
    \hfill
    \begin{subfigure}[b]{0.32\textwidth}
        \centering
        \includegraphics[width=\linewidth]{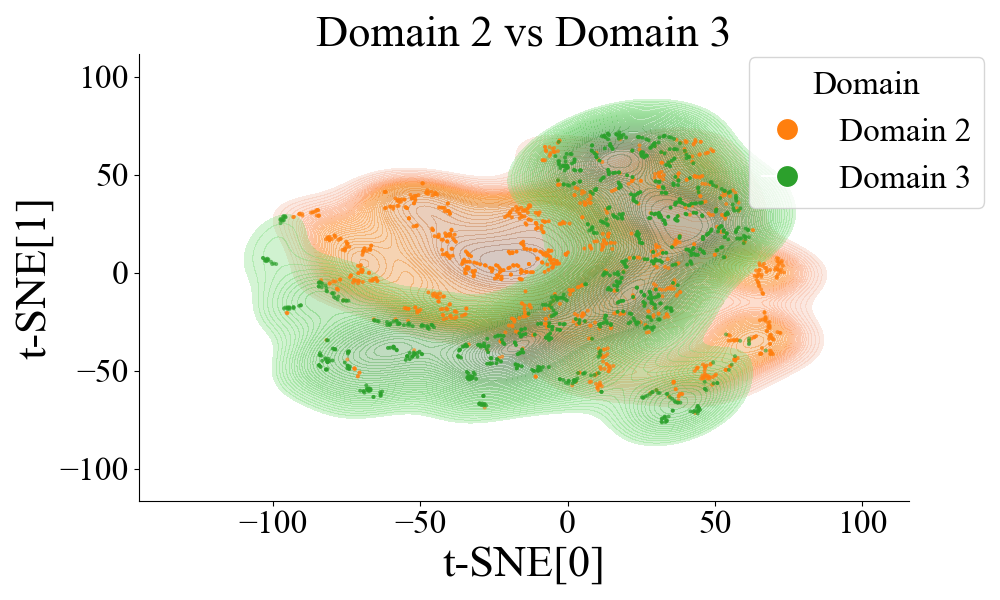}
        \caption{Domain 2 vs 3}
    \end{subfigure}
    \vspace{-0.5em}
    \caption{Pairwise t-SNE visualizations with density contours overlaid for each domain. Contour maps highlight structural differences in feature distributions across domains.}
    \label{fig:tsne_vis}
\end{figure*}

\paragraph{Unified training objective.}
The overall learning objective at incremental step $t$ is formulated as:
\begin{equation}
\mathcal{L}_{\text{total}}^{(t)}
=
\mathcal{L}_{\text{det}}^{(t)}
+
\mathcal{P}(\theta_t;\theta_{t-1})
=
\mathcal{L}_{\text{det}}^{(t)}
+
\lambda_f \, \mathcal{L}_{\text{feat}}
+
\lambda_h \, \mathcal{L}_{\text{head}}
\label{eq:total_loss}
\end{equation}
where $\mathcal{L}_{\text{det}}^{(t)}$ denotes the standard Faster R-CNN detection loss on the current domain.

This formulation enables end-to-end optimization using standard stochastic gradient descent, while implicitly approximating a constrained optimization process that suppresses accumulated deviation and stabilizes learning dynamics. Compared to explicit projection-based CL schemes, our penalty-based approach offers a more flexible and computationally efficient mechanism to control the optimization trajectory.

\paragraph{Optimization interpretation.}
From a geometric viewpoint, the preservation functional in Eq.~\eqref{eq:preserve_func} reshapes the optimization landscape by constraining updates within a structured low-drift subspace, thereby guiding the learning process toward solutions that remain transferable across domains. Under hierarchical sparsity, this implicit constraint induces a spectral regularization effect that improves the effective gradient signal-to-noise ratio and suppresses background-dominated perturbations. 

A rigorous theoretical analysis of this optimization behavior, including its spectral interpretation and its connection to accumulated deviation suppression, is provided in Appendix A-D. Algorithm~\ref{alg:training} summarizes the resulting training procedure.

\section{Dataset Analysis of Extreme Sparsity and Domain Shift}
\label{sec:dataset}
This section characterizes the statistical properties and optimization challenges of extremely sparse visual streams, utilizing the SpaceDet~\cite{xiao2025spacedet} dataset as a representative testbed. Section~\ref{sec:dataset_description} describes the dataset composition and domain partitioning. Section~\ref{sec:statistical_visual} presents statistical properties and visual analyses that highlight the challenges of object detection under extreme sparsity. Section~\ref{sec:domain_shift} focuses on feature-level domain shift analysis using quantitative metrics and t-SNE visualization.

\subsection{Dataset Description}
\label{sec:dataset_description}

The SpaceDet dataset~\cite{xiao2025spacedet} is a large-scale benchmark characterized by extreme visual sparsity and dynamic observation conditions. It consists of high-resolution grayscale imagery ($4418 \times 4418$ pixels, 16-bit) captured from multiple simulated sensors. Scenes contain multiple RSOs exhibiting diverse orbital altitudes, apparent sizes, aspect ratios, and illumination conditions, reflecting diverse observation conditions in real-world visual systems.

Fig.~\ref{fig:dataset_overview} illustrates representative characteristics. As shown in Fig.~\ref{fig:dataset_overview}(a), the raw 16-bit image appears almost empty due to extreme target sparsity. For visualization, Fig.~\ref{fig:dataset_overview}(b) presents a contrast-enhanced reference, revealing RSOs distributed across the field of view. Fig.~\ref{fig:dataset_overview}(c) zooms into sample regions, where blue, green, and red boxes indicate objects in Low Earth Orbit (LEO), Medium Earth Orbit (MEO), and Geostationary Earth Orbit (GEO), respectively. The statistical plots in Fig.~\ref{fig:dataset_overview}(d) confirm that most targets belong to the MEO class and occupy less than $0.005\%$ of the image area, underscoring the challenge of detecting tiny objects in a large field of view.

These properties, including extreme sparsity, large scale variations, and low signal-to-noise ratios compared to conventional datasets such as COCO~\cite{lin2014microsoft} or PASCAL VOC~\cite{everingham2010pascal}, make SpaceDet particularly challenging for CNN-based detectors prone to overfitting background features. This unique combination makes \textit{SpaceDet} a representative scenario for evaluating robust detection algorithms under domain shift.

\subsection{Statistical and Visual Characteristics} \label{sec:statistical_visual}

The \textit{SpaceDet} dataset exhibits distinctive properties presenting unique challenges for COD under extreme sparsity. A key challenge is extreme sparsity: RSO bounding boxes occupy less than 1\% of the total image area, with many spanning only a few pixels. To quantify signal quality, we compute the signal-to-noise ratio (SNR) as:
\begin{equation}
\mathrm{SNR} = 10 \cdot \log_{10} \left( \frac{\mu_{\mathrm{fg}}^2}{\sigma_{\mathrm{bg}}^2 + \epsilon} \right)
\label{eq:snr}
\end{equation}
where $\mu_{\mathrm{fg}}$ and $\sigma_{\mathrm{bg}}$ denote the average foreground intensity and background standard deviation, respectively. Based on this, the average SNR of \textit{SpaceDet} is measured as --0.87\,dB, significantly lower than standard benchmarks like COCO (6.26\,dB)~\cite{lin2014microsoft} and PASCAL VOC (5.25\,dB)~\cite{everingham2010pascal}. This highlights the extremely low signal strength of targets in our dataset.

Such sparse, low-SNR conditions heighten detection difficulty, promoting background overfitting~\cite{lin2017focal}. Additionally, the dataset exhibits pronounced spatiotemporal variations. Fig.~\ref{fig:target_density_curve} plots per-image target counts, revealing strong temporal fluctuations and inter-domain variation. Furthermore, Fig.~\ref{fig:target_heatmap} visualizes spatial distributions across the three sequential training subsets. Distinct density patterns are observed between domains, likely arising from changes in orbital geometry, illumination, and relative motion.

Overall, these characteristics—low SNR, extreme sparsity, and spatiotemporal shifts—expose the RSO detection challenge, motivating CL strategies adaptable to evolving observation conditions.

\subsection{Domain Shift Analysis}
\label{sec:domain_shift}

To analyze feature-level discrepancies, intermediate features $\{\mathbf{f}_i\}_{i=1}^{N}$ are extracted from target patches using a trained backbone, where $\mathbf{f}_i \in \mathbb{R}^D$. To improve efficiency, we apply Principal Component Analysis (PCA) to project features into a 50-dimensional subspace $\mathbf{z}_i = \mathbf{W}_{\text{PCA}}^\top \mathbf{f}_i$.

Subsequently, t-SNE embeds $\{\mathbf{z}_i\}$ into a 2D space by minimizing the Kullback-Leibler divergence:
\begin{equation}
\min_{\{\mathbf{y}_i\}} \mathrm{KL}(P \,||\, Q) = \sum_{i \neq j} p_{ij} \log \frac{p_{ij}}{q_{ij}}
\end{equation}
where $p_{ij}$ and $q_{ij}$ denote affinities in the original and embedded spaces, respectively. Auxiliary contextual information is appended to $\mathbf{z}_i$ prior to projection to enhance separation under sparse conditions.

As shown in Fig.~\ref{fig:tsne_vis}, the embeddings $\{\mathbf{y}_i\}$ exhibit clear clustering between domains, indicating substantial distributional shift despite consistent semantics. This validates the necessity of CL strategies to handle inter-domain variations.

To quantify shift across domains, three pairwise metrics are computed over feature sets $\{\mathbf{f}_i^{(k)}\}$: Maximum Mean Discrepancy (MMD), Wasserstein distance ($\mathrm{D}_{\mathrm{W}}$), and domain classification accuracy (DA) derived from a binary classifier trained on $\mathbf{f}_i \in \mathcal{F}_1 \cup \mathcal{F}_2$.

\begin{table}[H]
\centering
\renewcommand{\arraystretch}{1.15}
\setlength{\tabcolsep}{8pt}
\caption{Pairwise domain discrepancy metrics. DA denotes domain classification accuracy.}
\label{tab:domain_shift}
\begin{tabular}{lccc}
\toprule
\textbf{Domain Pair} & \textbf{MMD} & \textbf{Wasserstein} & \textbf{DA} (\%) \\
\midrule
Domain 1 vs 2  & 0.088 & 0.018 & 98.0 \\
Domain 1 vs 3  & 0.259 & 0.032 & 97.3 \\
Domain 2 vs 3  & 0.050 & 0.011 & 97.3 \\
\bottomrule
\end{tabular}
\vspace{0.5em}
\end{table}

As shown in Table~\ref{tab:domain_shift}, all domain pairs exhibit relatively low MMD (ranging from 0.05 to 0.26) and Wasserstein distance (from 0.011 to 0.032), indicating moderate distributional divergence. However, the domain classifier consistently achieves high classification accuracy (DA $\geq$ 97.3\%), suggesting that despite global similarity, there exist task-relevant structural differences in the learned features. This discrepancy further emphasizes the necessity of domain-adaptive CL strategies.

\begin{table*}[t]
\centering
\setlength{\tabcolsep}{5pt}
\renewcommand{\arraystretch}{1.25}
\caption{Comparison with existing detectors and CL methods under a domain-incremental setting.
“Domain~1” and “Domain~2” are the two sequential domains. “Total Metric” is the macro-average over the two domains. mAP reports mAP@50. Pre., Rec., and F1 denote precision, recall, and F1-score respectively.}
\label{tab:main_results}
\begin{tabular}{
    l|
    >{\columncolor{myblue}}c >{\columncolor{myblue}}c >{\columncolor{myblue}}c >{\columncolor{myblue}}c |
    >{\columncolor{mygreen}}c >{\columncolor{mygreen}}c >{\columncolor{mygreen}}c >{\columncolor{mygreen}}c |
    >{\columncolor{myyellow}}c >{\columncolor{myyellow}}c >{\columncolor{myyellow}}c >{\columncolor{myyellow}}c
}
\toprule
\multirow{2}{*}{\textbf{Method}}
& \multicolumn{4}{c|}{\textbf{Domain 1}}
& \multicolumn{4}{c|}{\textbf{Domain 2}}
& \multicolumn{4}{c}{\textbf{Total Metric}} \\
& \textbf{Pre.} & \textbf{Rec.} & \textbf{F1} & \textbf{mAP}
& \textbf{Pre.} & \textbf{Rec.} & \textbf{F1} & \textbf{mAP}
& \textbf{Pre.} & \textbf{Rec.} & \textbf{F1} & \textbf{mAP} \\
\midrule
Single Domain 1       & \textbf{49.38} & 47.00 & 48.16 & \textbf{47.86}  & 41.85 & 41.00 & 41.42 & 31.52  & 45.61 & 44.00 & 44.79 & 39.69 \\
Single Domain 2       & 31.69 & 33.00 & 32.33 & 28.98  & 48.98 & 44.00 & 46.36 & 36.32  & 40.34 & 38.50 & 39.35 & 32.65 \\
Sequential w/o CL     & 37.92 & 37.00 & 37.46 & 34.28  & 44.25 & 47.00 & 45.59 & 36.18  & 41.09 & 42.00 & 41.52 & 35.23 \\
Joint Training         & 48.42 & \textbf{48.00} & \textbf{48.21} & 47.05  & 52.06 & 45.00 & 48.27 & 38.06  & 50.24 & 46.50 & 48.24 & 42.56 \\
\midrule
YOLOv10m\cite{wang2024yolov10}
& 40.39 & 38.00 & 39.15 & 37.71
& 48.54 & 46.00 & 47.23 & 36.58
& 44.47 & 42.00 & 43.19 & 37.14 \\
Faster R-CNN\cite{ren2015faster}
& 37.92 & 37.00 & 37.46 & 34.28
& 44.25 & 47.00 & 45.59 & 36.18
& 41.09 & 42.00 & 41.52 & 35.23 \\
DETR\cite{carion2020end}
& 6.54 & 10.00 & 7.91 & 1.59
& 9.21 & 16.00 & 11.69 & 2.32
& 7.87 & 13.00 & 9.80 & 1.81 \\
Deformable DETR\cite{zhu2020deformable}
& 38.62 & 39.00 & 38.81 & 36.89
& 53.87 & 44.00 & 48.44 & 40.31
& 46.24 & 41.50 & 43.62 & 38.60 \\
\midrule
Shmelkov et al.\cite{shmelkov2017incremental}
& 42.57 & 45.00 & 43.75 & 37.84
& 52.54 & 44.00 & 47.89 & 36.78
& 47.56 & 44.50 & 45.82 & 37.31 \\
EWC\cite{kirkpatrick2017overcoming}
& 50.09 & 32.62 & 39.51 & 21.31
& 76.44 & 36.79 & 49.67 & \textbf{46.69}
& \textbf{63.27} & 34.71 & 44.59 & 33.99 \\
OGD\cite{farajtabar2020orthogonal}
& 41.81 & 29.23 & 34.41 & 19.09
& \textbf{83.48} & 34.14 & 48.46 & 44.62
& 62.65 & 31.69 & 41.43 & 31.85 \\
CWD\cite{liang2023loss}
& 42.25 & 27.06 & 32.99 & 22.31
& 78.51 & 30.40 & 43.83 & 44.18
& 60.38 & 28.73 & 38.41 & 33.24 \\
CL-DETR\cite{liu2023continual}
& 26.72 & 13.00 & 17.49 & 9.28
& 24.50 & 24.00 & 24.25 & 7.14
& 25.61 & 18.50 & 20.87 & 8.21 \\
\rowcolor{myorange}
\textbf{Ours}
& 44.99 & 47.00 & 45.97 & 44.22
& 53.73 & \textbf{48.00} & \textbf{50.71} & 41.03
& 49.36 & \textbf{47.50} & \textbf{48.34} & \textbf{42.62} \\
\bottomrule
\end{tabular}
\vspace{0.5em}
\end{table*}

\begin{table*}[t]
\centering
\setlength{\tabcolsep}{4.5pt}
\renewcommand{\arraystretch}{1.25}
\caption{
Ablation study on different components of the proposed framework. 
``Patch'' denotes patch selection, ``Aug'' denotes domain-aware augmentation, 
``Feat'' and ``Head'' represent feature-level and head-level distillation, respectively.
}
\label{tab:ablation}
\begin{tabular}{
    c c c c |
    >{\columncolor{myblue}}c >{\columncolor{myblue}}c >{\columncolor{myblue}}c >{\columncolor{myblue}}c |
    >{\columncolor{mygreen}}c >{\columncolor{mygreen}}c >{\columncolor{mygreen}}c >{\columncolor{mygreen}}c |
    >{\columncolor{myyellow}}c >{\columncolor{myyellow}}c >{\columncolor{myyellow}}c >{\columncolor{myyellow}}c
}
\toprule
\multirow{2}{*}{\textbf{Patch}} 
& \multirow{2}{*}{\textbf{Aug}} 
& \multirow{2}{*}{\textbf{Feat}} 
& \multirow{2}{*}{\textbf{Head}} 
& \multicolumn{4}{c|}{\textbf{Task 1}} 
& \multicolumn{4}{c|}{\textbf{Task 2}} 
& \multicolumn{4}{c}{\textbf{Total Metric}} \\
& & & 
& \textbf{Pre.} & \textbf{Rec.} & \textbf{F1} & \textbf{mAP}
& \textbf{Pre.} & \textbf{Rec.} & \textbf{F1} & \textbf{mAP}
& \textbf{Pre.} & \textbf{Rec.} & \textbf{F1} & \textbf{mAP} \\
\midrule
$\times$ & $\times$ & $\times$ & $\times$
& 0.02 & 0.04 & 0.03 & 0.00
& 0.02 & 0.04 & 0.03 & 0.00
& 0.02 & 0.04 & 0.03 & 0.00 \\

\checkmark & $\times$ & $\times$ & $\times$
& 28.32 & 26.66 & 26.38 & 18.38
& 20.93 & 47.41 & 27.94 & 37.03
& 24.63 & 37.03 & 27.16 & 27.70 \\

\checkmark & \checkmark & $\times$ & $\times$
& 37.92 & 37.00 & 37.46 & 34.28
& 44.25 & 47.00 & 45.59 & 36.18
& 41.09 & 42.00 & 41.52 & 35.23 \\

\checkmark & \checkmark & $\times$ & \checkmark
& 42.57 & 45.00 & 43.75 & 37.84
& 52.54 & 44.00 & 47.89 & 36.78
& 47.56 & 44.50 & 45.82 & 37.31 \\

\checkmark & \checkmark & \checkmark & $\times$
& 42.82 & 36.29 & 39.28 & 30.74
& 45.74 & 42.70 & 44.17 & \textbf{42.47}
& 44.28 & 39.50 & 41.73 & 36.61 \\

\rowcolor{myorange}
\checkmark & \checkmark & \checkmark & \checkmark
& \textbf{44.99} & \textbf{47.00} & \textbf{45.97} & \textbf{44.22}
& \textbf{53.73} & \textbf{48.00} & \textbf{50.71} & 41.03
& \textbf{49.36} & \textbf{47.50} & \textbf{48.34} & \textbf{42.62} \\
\bottomrule
\end{tabular}
\vspace{0.5em}
\end{table*}

\section{Experiments and Results}
\label{sec:experiments}

We evaluate our proposed COD approach under a domain-incremental setup using the SpaceDet dataset~\cite{xiao2025spacedet}. To establish a comprehensive benchmark, we compare our method against a diverse set of baselines, categorized into standard training paradigms (e.g., sequential fine-tuning), mainstream object detectors, and representative replay-free CL approaches (e.g., Shmelkov et al.~\cite{shmelkov2017incremental}, CL-DETR~\cite{liu2023continual}). The following sections detail the datasets, experimental configurations, implementation specifics, and empirical results.

\subsection{Datasets and Evaluation Metrics}

We evaluate our framework on the SpaceDet dataset~\cite{xiao2025spacedet}. To align with the temporally evolving space object detection tasks, we select a continuous sequence of 1500 images from one of the four cameras (Camera~2) as a representative example. Specifically, the first 500 images are assigned to $\mathcal{D}^{(1)}$, the next 500 to $\mathcal{D}^{(2)}$, and the final 500 to $\mathcal{D}^{(3)}$, enabling evaluation on a third sequential domain. Each domain is split into training, validation, and test subsets using a 7:2:1 ratio. This design preserves temporal continuity while enabling evaluation under naturally varying observational conditions. Importantly, our setting extends to sequences from any camera within \textit{SpaceDet} where domain shift is present.

Average Precision at 0.5 IoU (AP50) is used as the primary evaluation metric. To assess inter-domain distributional shift, we additionally report Maximum Mean Discrepancy (MMD), Wasserstein distance, and domain classification accuracy, which are widely adopted metrics for characterizing domain divergence.

\subsection{Experimental Settings} \label{sec:experimental_settings}

We simulate a COD scenario under domain shift by partitioning the dataset into three temporally ordered domains: $\mathcal{D}^{(1)}$, $\mathcal{D}^{(2)}$, and $\mathcal{D}^{(3)}$. Each domain reflects a distinct input distribution $P^{(t)}(X, Y)$ arising from evolving observation geometry and illumination. In the primary protocol, the detector is trained on $\mathcal{D}^{(1)}$ and continually adapted to $\mathcal{D}^{(2)}$ without accessing previous data. We also evaluate the reverse order $\mathcal{D}^{(2)} \rightarrow \mathcal{D}^{(1)}$ to ensure robustness against order bias. To further assess scalability, a third domain $\mathcal{D}^{(3)}$ is introduced in extended experiments.

To evaluate our framework, we compare it against mainstream detectors and representative CL baselines, including Shmelkov et al.~\cite{shmelkov2017incremental}, CL-DETR~\cite{liu2023continual}, and classical methods like EWC~\cite{kirkpatrick2017overcoming}, OGD~\cite{farajtabar2020orthogonal}, and CWD~\cite{liang2023loss}. All methods use identical conditioning to ensure fairness. We also conduct ablation studies to analyze individual components and utilize $\mathcal{D}^{(3)}$ to assess long-term adaptation. Additionally, we include standard references: (a) \textit{single-domain training} on $\mathcal{D}^{(1)}$ or $\mathcal{D}^{(2)}$; (b) \textit{joint training} (upper bound) merging both domains; and (c) \textit{sequential fine-tuning} without CL mechanisms. All methods are evaluated under both orders, with detailed results reported in Sec.~\ref{sec:results}.

\subsection{Implementation Details}

Our COD framework is built on Faster R-CNN~\cite{ren2015faster}, which integrates an RPN with a two-branch detection head. To mitigate catastrophic forgetting when adapting to a new domain $\mathcal{D}^{(t)}$, we keep a frozen copy of the previous detector $\theta_{t-1}$ and transfer its knowledge to the updated model $\theta_t$ via dual-stage preservation constraints, which regularize the update direction to maintain consistency on both backbone features and RoI-level outputs.

\begin{table*}[t]
\centering
\setlength{\tabcolsep}{4.5pt}
\renewcommand{\arraystretch}{1.25}
\caption{Comparison across three sequential domains. Each task corresponds to a distinct domain shift.}
\label{tab:three_domain_results}
\begin{tabular}{
    l|
    >{\columncolor{myblue}}c >{\columncolor{myblue}}c >{\columncolor{myblue}}c >{\columncolor{myblue}}c |
    >{\columncolor{mygreen}}c >{\columncolor{mygreen}}c >{\columncolor{mygreen}}c >{\columncolor{mygreen}}c |
    >{\columncolor{mypink}}c >{\columncolor{mypink}}c >{\columncolor{mypink}}c >{\columncolor{mypink}}c |
    >{\columncolor{myyellow}}c >{\columncolor{myyellow}}c >{\columncolor{myyellow}}c >{\columncolor{myyellow}}c
}
\toprule
\multirow{2}{*}{\textbf{Method}} 
& \multicolumn{4}{c|}{\textbf{Domain 1}} 
& \multicolumn{4}{c|}{\textbf{Domain 2}} 
& \multicolumn{4}{c|}{\textbf{Domain 3}} 
& \multicolumn{4}{c}{\textbf{Total Metric}} \\
& \textbf{Pre.} & \textbf{Rec.} & \textbf{F1} & \textbf{mAP} 
& \textbf{Pre.} & \textbf{Rec.} & \textbf{F1} & \textbf{mAP} 
& \textbf{Pre.} & \textbf{Rec.} & \textbf{F1} & \textbf{mAP} 
& \textbf{Pre.} & \textbf{Rec.} & \textbf{F1} & \textbf{mAP} \\
\midrule
Single Domain 1     & \textbf{49.38} & 47.00 & 48.16 & \textbf{47.86} & 41.85 & 41.00 & 41.42 & 31.52 & 25.93 & 31.00 & 28.24 & 18.35 & 39.05 & 39.67 & 39.27 & 32.58 \\
Single Domain 2     & 31.69 & 33.00 & 32.33 & 28.98 & 48.98 & 44.00 & 46.36 & 36.32 & 38.93 & 38.00 & 38.46 & 21.30 & 39.87 & 38.33 & 39.05 & 28.87 \\
Single Domain 3     & 25.12 & 14.00 & 17.98 & 14.18 & 21.23 & 21.00 & 21.12 & 8.40  & \textbf{59.93} & \textbf{55.00} & \textbf{57.36} & 38.05 & 35.43 & 30.00 & 32.15 & 20.21 \\
Sequential w/o CL   & 32.54 & 22.00 & 26.25 & 25.25 & 28.14 & 23.00 & 25.31 & 15.56 & 56.10 & 50.00 & 52.87 & \textbf{38.88} & 39.92 & 31.67 & 34.81 & 26.56 \\
Joint Training      & 47.61 & 48.00 & 47.81 & 46.44 & 47.83 & \textbf{47.00} & 47.41 & 35.18 & 54.15 & 44.00 & 48.55 & 32.63 & 49.86 & 46.33 & 47.92 & 38.08 \\
\rowcolor{myorange}
\textbf{Ours}        & 48.42 & \textbf{48.00} & \textbf{48.21} & 47.05 & \textbf{52.06} & 45.00 & \textbf{48.27} & \textbf{39.06} & 57.69 & 50.00 & 53.57 & 36.26 & \textbf{52.72} & \textbf{48.67} & \textbf{50.02} & \textbf{40.79} \\
\bottomrule
\end{tabular}
\vspace{0.5em}
\end{table*}

\begin{table*}[t]
\centering
\setlength{\tabcolsep}{5pt}
\renewcommand{\arraystretch}{1.25}
\caption{Evaluation under reversed domain sequence ($\mathcal{D}^{(2)}$ $\rightarrow$ $\mathcal{D}^{(1)}$).}
\label{tab:order_reversal}
\begin{tabular}{
    l|
    >{\columncolor{myblue}}c >{\columncolor{myblue}}c >{\columncolor{myblue}}c | >{\columncolor{myblue}}c |
    >{\columncolor{mygreen}}c >{\columncolor{mygreen}}c >{\columncolor{mygreen}}c | >{\columncolor{mygreen}}c |
    >{\columncolor{myyellow}}c >{\columncolor{myyellow}}c >{\columncolor{myyellow}}c | >{\columncolor{myyellow}}c
}
\toprule
\multirow{2}{*}{\textbf{Method}} 
& \multicolumn{4}{c|}{\textbf{Domain 1}} 
& \multicolumn{4}{c|}{\textbf{Domain 2}} 
& \multicolumn{4}{c}{\textbf{Total Metric}} \\
& \textbf{Pre.} & \textbf{Rec.} & \textbf{F1} & \textbf{mAP} 
& \textbf{Pre.} & \textbf{Rec.} & \textbf{F1} & \textbf{mAP} 
& \textbf{Pre.} & \textbf{Rec.} & \textbf{F1} & \textbf{mAP} \\
\midrule
Single Domain 1   & 49.38 & 47.00 & 48.16 & 47.86 & 41.85 & 41.00 & 41.42 & 31.52 & 45.61 & 44.00 & 44.79 & 39.69 \\
Single Domain 2   & 31.69 & 33.00 & 32.33 & 28.98 & 48.98 & 44.00 & 46.36 & 36.32 & 40.34 & 38.50 & 39.35 & 32.65 \\
Sequential w/o CL & 43.10 & 47.00 & 44.97 & 44.90 & 50.70 & 40.00 & 44.72 & 34.81 & 46.90 & 43.50 & 44.84 & 39.85 \\
Joint Training    & 48.42 & \textbf{48.00} & \textbf{48.21} & 47.05 & 52.06 & \textbf{45.00} & \textbf{48.27} & \textbf{38.06} & 50.24 & \textbf{46.50} & \textbf{48.24} & \textbf{42.56} \\
\rowcolor{myorange}
\textbf{Ours}     & \textbf{49.70} & 43.00 & 46.11 & \textbf{48.25} & \textbf{52.82} & 40.00 & 45.52 & 36.16 & \textbf{51.26} & 41.50 & 45.82 & 42.20 \\
\bottomrule
\end{tabular}
\vspace{0.5em}
\end{table*}

\begin{figure*}[t]
    \centering
    \includegraphics[width=0.90\linewidth]{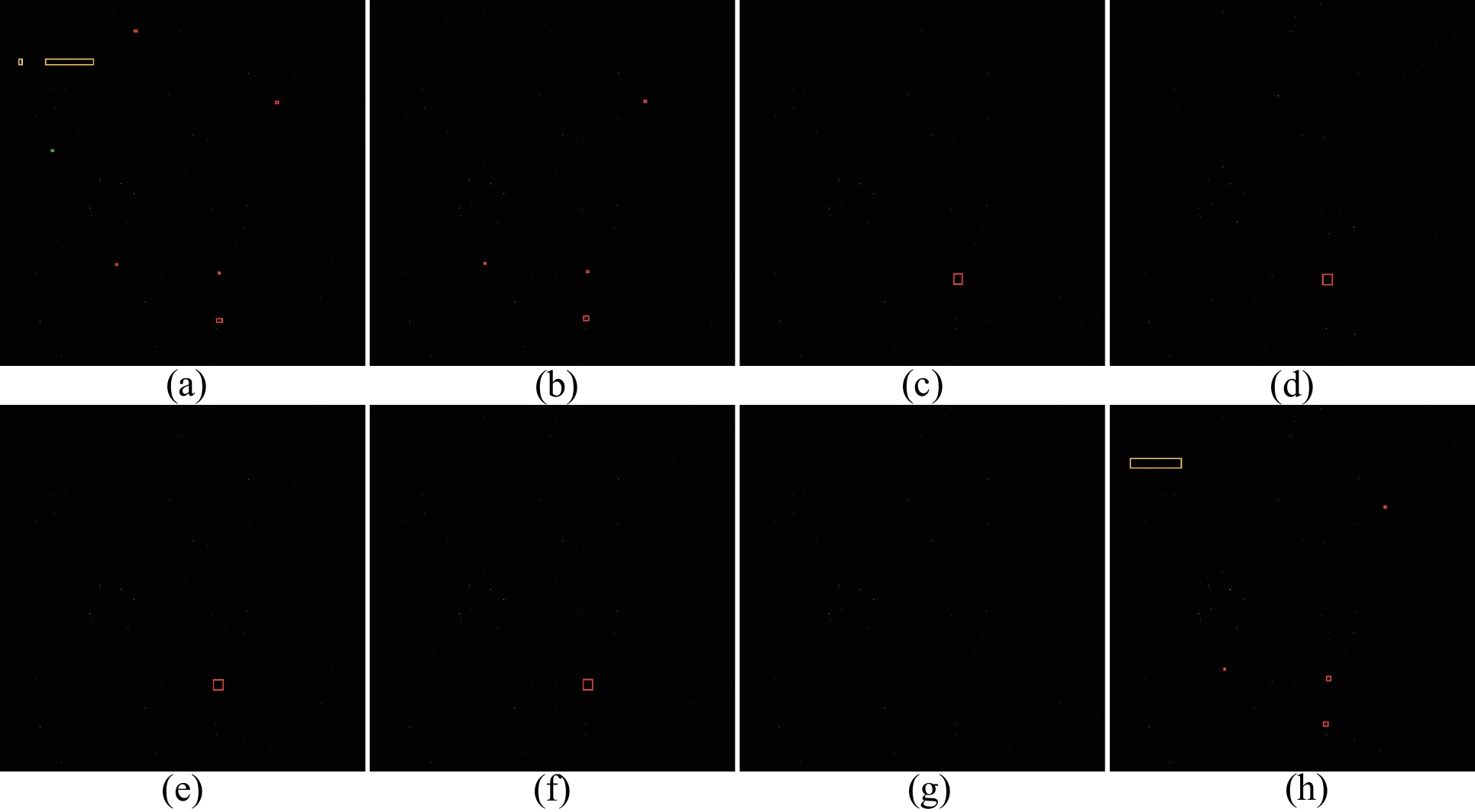}
    \caption{Qualitative comparison of detection results under the domain-incremental setting. 
    (a) Ground truth. 
    (b) Shmelkov et al.~\cite{shmelkov2017incremental}. 
    (c) Backbone-only distillation. 
    (d) EWC~\cite{kirkpatrick2017overcoming}. 
    (e) OGD~\cite{farajtabar2020orthogonal}. 
    (f) CWD~\cite{liang2023loss}. 
    (g) CL-DETR~\cite{liu2023continual}. 
    (h) Ours.}
    \label{fig:qualitative}
\end{figure*}

During training, the model is optimized under the standard detection objective $\mathcal{L}_{\text{det}}$, while its parameter updates are further guided by the preservation field (see Sec.~\ref{sec:framework}). Batch normalization layers in the backbone are frozen to preserve domain-invariant statistics, while the RPN and RoI head remain adaptable. The distillation objective provides regularization to stabilize learning under domain shift.

We adopt stochastic gradient descent (SGD) with momentum 0.9 as the optimizer. The initial learning rate is set to 0.02 and decayed to 0.0002 following a step-based schedule, with a linear warm-up for the first 100 iterations. For all experiments, the model is trained on a single A6000 GPU with a batch size of 16.

Each domain is trained for 25{,}000 iterations. In the continual setting, the model is trained on $\mathcal{D}^{(1)}$ for 18{,}000 iterations, then on $\mathcal{D}^{(2)}$ for 7{,}000. For baselines, we train single-domain models and joint-domain models for 25{,}000 iterations. The sequential fine-tuning baseline also follows the 18k+7k training schedule.

During inference, we retain up to 100 predictions per image after applying non-maximum suppression (NMS) with an IoU threshold of 0.4. Our implementation is built upon the Detectron2 \cite{wu2019detectron2} library.

\subsection{Results} \label{sec:results}
We organize our experimental evaluation into three main parts. First, we benchmark our method against a wide range of representative CL approaches, alongside standard training paradigms (e.g., Joint Training, Sequential Fine-tuning) to establish comprehensive upper and lower bounds (Sec.~\ref{sec:main_results}). Second, we verify the contribution of individual components through detailed ablation studies (Sec.~\ref{sec:ablation}). Finally, we evaluate the scalability and robustness of the framework by extending the scope to a three-domain setting and assessing its sensitivity to task order reversal.

\subsubsection{Comparative Evaluation against Representative Baselines}
\label{sec:main_results}

To evaluate the effectiveness of our framework under domain-incremental settings, we benchmark our method against a diverse set of baselines, including mainstream object detectors, representative replay-free CL approaches, and standard training paradigms. All models are trained sequentially from $\mathcal{D}^{(1)}$ to $\mathcal{D}^{(2)}$ using the same conditioning pipeline to ensure a fair comparison.

As summarized in Table~\ref{tab:main_results}, conventional detectors and the \textit{Sequential w/o CL} baseline exhibit limited robustness, leading to noticeable performance degradation on the previous domain (e.g., mAP drops to 34.28\%). While \textit{Joint Training} maintains balanced cross-domain performance (42.56\% total mAP), such an idealized setting is often impractical. Classical replay-free CL methods, including EWC, OGD, and CWD, demonstrate stronger plasticity to the new domain (e.g., EWC reaches 46.69\%) but incur severe forgetting on the previous domain (dropping below 22\%), resulting in relatively low overall accuracy ($\approx 34\%$). This behavior reflects the intrinsic challenge of hierarchical sparsity, where background-dominated gradients bias importance estimation, hindering stable knowledge preservation (see Appendix A-A). Compared with these baselines, Shmelkov et al. achieves more balanced performance (37.31\% mAP), yet its head-only constraint limits further improvement. In contrast, our dual-stage framework attains the highest total mAP of 42.62\%, even slightly surpassing the \textit{Joint Training} baseline (42.56\%). We attribute this to the regularization effect of the dual-stage invariance, which mitigates overfitting to dominant background patterns in joint training. Fig.~\ref{fig:qualitative} further visualizes that our method detects substantially more tiny RSOs with accurate localization, while most baseline approaches suffer from severe miss detections.

\subsubsection{Ablation Study and Component Analysis}
\label{sec:ablation}

We conduct ablation experiments to analyze the contribution of each component, including patch selection, domain-aware augmentation, and dual-stage invariance. All variants are evaluated under the same domain-incremental setting, with results summarized in Table~\ref{tab:ablation}.

Without conditioning or distillation, sequential training collapses due to severe background dominance, yielding near-zero performance. Introducing patch selection substantially raises total mAP to 27.70, verifying its necessity in constructing target-centric proposals and suppressing noise. Domain-aware augmentation further boosts mAP to 35.23 by alleviating imbalance and enhancing cross-domain generalization. On top of conditioning, head-only distillation improves semantic stability (37.31 mAP) yet fails to sufficiently constrain backbone drift. Conversely, feature-only distillation regularizes backbone representations but suffers from increased forgetting, resulting in inferior domain balance (36.61 mAP). By jointly enforcing consistency, dual-stage invariance achieves the best performance (42.62 mAP), validating its efficacy in suppressing accumulated deviation and maintaining a balanced stability--plasticity trade-off.

\subsubsection{Generalization to Additional Domains}
To evaluate the generalization capability of the proposed COD framework under an expanded domain scenario, a third domain $\mathcal{D}^{(3)}$ is introduced. In the sequential configuration, the model is trained on $\mathcal{D}^{(1)}$, adapted to $\mathcal{D}^{(2)}$, and further adapted to $\mathcal{D}^{(3)}$ without revisiting previous domains. All evaluations are conducted on each individual domain to assess whether the model can maintain consistent detection performance and mitigate catastrophic forgetting as the number of domains increases.

Table~\ref{tab:three_domain_results} reports the Precision, Recall, F1-score, and mAP@50 across all three tasks. In the \textit{Single Domain} setting, models perform well only on their own domain, confirming the impact of domain shift. The \textit{Sequential w/o CL} configuration further degrades performance, showing that training on additional domains without adaptation exacerbates forgetting. \textit{Joint Training}, which leverages all domains simultaneously, achieves relatively strong results. However, as more domains with increasing distributional divergence are introduced, its overall performance declines compared to the two-domain case. In contrast, the proposed \textit{Ours} framework outperforms all compared paradigms across most evaluation metrics, demonstrating its superior effectiveness in mitigating domain shifts. In addition, it maintains balanced accuracy across all tasks, highlighting its robustness and generalization capability under domain-incremental scenarios.

Notably, while our method exhibits a slight performance drop on $\mathcal{D}^{(2)}$ due to the need to accommodate $\mathcal{D}^{(3)}$, it achieves substantial improvements on $\mathcal{D}^{(1)}$. This suggests that the similarity between certain domains can facilitate knowledge transfer and enhance generalization. As more domains are incorporated, the overall detection performance tends to degrade due to the increasing distributional discrepancies, which further emphasizes the importance of effective CL mechanisms for maintaining cross-domain robustness.

\subsubsection{Robustness to Task Sequence Variations}
To test order sensitivity, the training sequence is reversed from $\mathcal{D}^{(2)}$ to $\mathcal{D}^{(1)}$. As shown in Table~\ref{tab:order_reversal}, all settings show similar trends as before. In particular, the \textit{Sequential w/o CL} baseline suffers from notable forgetting, while the \textit{Ours} configuration consistently maintains balanced performance across domains.  The total mAP remains close to the forward-order case, confirming the robustness of our method to task sequence variations.

Task sequence plays a crucial role in CL, where earlier domains are more susceptible to forgetting. To assess the order sensitivity of our approach, we reverse the training sequence from $\mathcal{D}^{(1)} \rightarrow \mathcal{D}^{(2)}$ to $\mathcal{D}^{(2)} \rightarrow \mathcal{D}^{(1)}$. As shown in Table~\ref{tab:order_reversal}, all configurations exhibit similar performance trends as in the original setting. In particular, the \textit{Sequential w/o CL} baseline suffers from notable forgetting on Task~1, while the \textit{Ours} configuration consistently maintains balanced detection across tasks. The total mAP remains comparable to the forward-order results, confirming that the proposed method is robust to task sequence variations. Importantly, this robustness is achieved without any additional tuning, underscoring the stability and generalizability of our framework.

\section{Conclusion}
\label{sec:conclusion}
Existing COD methods often struggle with domain shift and performance degradation in realistic domain-incremental scenarios, particularly in complex settings such as high-resolution space-based imagery. In this work, we propose a dual-stage invariant framework for COD. It explicitly suppresses accumulated representation shift induced by hierarchical sparsity through joint constraints on backbone features and detection predictions, and is further supported by a sparsity-aware data conditioning pipeline.
Our method effectively mitigates catastrophic forgetting while preserving adaptability to new domains. Extensive experiments across three sequential domains demonstrate that the proposed approach achieves performance close to joint training upper bounds, and consistently outperforms standard sequential baselines. The framework remains stable as the number of domains increases, exhibiting strong robustness to domain shifts and scalability to more realistic continual detection scenarios.

A key future direction lies in enabling online CL~\cite{nie2023online}, which is essential for deployment in long-term autonomous systems~\cite{lin2025multi}. Notably, the sequential and memory-free nature of our framework makes it well-suited for online training scenarios. Beyond this, multi-modal sensor fusion for space situational awareness and extensions to related tasks such as instance segmentation constitute promising directions for future research.

\bibliographystyle{IEEEtran}
\bibliography{bibtex/bib/IEEEabrv,bibtex/bib/myrefs}
\vfill
\clearpage

\appendices
\section{Theoretical Analysis of Gradient Pathology}
\label{app:theoretical_analysis}
This appendix presents a theoretical study of gradient pathology in space-based tiny object detection under domain-incremental learning. We analyze the degradation of gradient propagation caused by extreme foreground sparsity and domain shift, and formally justify the necessity of dual-stage invariance for stabilizing optimization and mitigating catastrophic forgetting.

\subsection{Hierarchical Sparsity and Gradient Dominance}
\label{app:nested_sparsity}

This appendix formalizes the optimization characteristics of space-based tiny object detection by modeling the hierarchical sparsity property. We show that due to extreme inter- and intra-proposal sparsity, the effective gradient signal is dominated by background components.

\vspace{2pt}
\noindent 1) Proposal-level Objective and Gradient Decomposition:
Let $x$ be an input image. A two-stage detector generates candidate proposals $\mathcal{A}(x) = \{a_1,\ldots,a_{N(x)}\}$. The training loss and gradient are additive over proposals: $\mathcal{L}(\theta; x) = \sum_{a \in \mathcal{A}(x)} \ell(\theta; x, a)$ and $g(\theta; x) = \sum_{a \in \mathcal{A}(x)} g_a$.
We define positive/negative proposal sets as $\mathcal{A}^+(x) = \{ a \mid y_a=1 \}$ and $\mathcal{A}^-(x) = \{ a \mid y_a=0 \}$.
Let $\gamma(x) = |\mathcal{A}^+|/|\mathcal{A}|$ denote the positive density. In space-based detection, $\gamma(x)\ll 1$ holds. Accordingly, the total gradient decomposes as:
\begin{equation}
g(\theta; x) \;=\; g^{+} + g^{-} \;=\; \sum_{a\in \mathcal{A}^+(x)} g_a \;+\; \sum_{a\in \mathcal{A}^-(x)} g_a
\label{eq:grad_pos_neg_split}
\end{equation}

\vspace{2pt}
\noindent 2) Intra-proposal Sparsity and Gradient Dominance:
For a positive proposal $a\in\mathcal{A}^+$, let $\Omega_a$ denote its spatial support. The target occupancy ratio is $\nu_a = (\sum_{\Omega_a} M_{ij}) / |\Omega_a|$, where $M(x)$ is the binary mask. In space imagery, $\nu_a\ll 1$ typically holds.
The proposal loss and gradient decompose into object and background terms: $\ell = \ell^{obj} + \ell^{bg}$ and $g_a = g_a^{obj} + g_a^{bg}$. We make the following assumption on gradient energy:
\begin{equation}
\mathbb{E}\big[\|g_a^{obj}\|\big] \le C_{obj}\,\nu_a,
\quad
\mathbb{E}\big[\|g_a^{bg}\|\big] \ge C_{bg}\,(1-\nu_a)
\label{eq:occupancy_assumption}
\end{equation}
\emph{Remark:} This serves as a valid first-order proxy for gradient flow, reflecting the spatial aggregation mechanism where gradient contributions scale with active spatial support.
Combining Eq.~\eqref{eq:occupancy_assumption} yields the bound on relative gradient strength:
\begin{equation}
\frac{\mathbb{E}\big[\|g_a^{obj}\|\big]}
{\mathbb{E}\big[\|g_a^{bg}\|\big]}
\;\le\;
\frac{C_{obj}}{C_{bg}} \cdot \frac{\nu_a}{1-\nu_a}
\;\xrightarrow{\nu_a \to 0}\; 0
\label{eq:snr_bound_step1}
\end{equation}
When $\nu_a\ll 1$, the ratio vanishes, implying that even positive proposals provide background-dominated signals. This constitutes a fundamental challenge for continual optimization.

\subsection{Gradient SNR Collapse}
\label{app:snr_collapse}

Building upon Appendix~\ref{app:nested_sparsity}, we quantify optimization difficulty using the proposal-level \emph{gradient signal-to-noise ratio}: $\mathrm{SNR}_a \triangleq \mathbb{E}[\|g_a^{obj}\|] / \sqrt{\mathbb{E}[\|g_a^{bg}\|^2]}$.
By applying Jensen's inequality to the denominator and substituting the energy bound from Eq.~\eqref{eq:snr_bound_step1}, we obtain a tight upper bound:
\begin{equation}
\mathrm{SNR}_a
\;\le\;
\frac{C_{obj}}{C_{bg}} \left( \frac{\nu_a}{1-\nu_a} \right)
\label{eq:snr_collapse_final}
\end{equation}
Under extreme sparsity ($\nu_a \to 0$), this bound asymptotically vanishes. This formally proves that valid optimization signals are numerically overwhelmed by background noise, necessitating the proposed dual-stage invariance.

\subsection{Spectral Bias of Fisher Information under Hierarchical Sparsity}
\label{app:fisher_bias}

Standard regularizers like EWC rely on the Fisher Information Matrix (FIM). Under hierarchical sparsity, gradients are background-dominated ($g \approx g^{bg}$), causing the FIM to collapse into the covariance of background statistics:
\begin{equation}
\mathbf{F} 
\;\approx\; 
\mathbb{E} \!\left[ g^{bg}(x)\, g^{bg}(x)^\top \right] 
\;+\; \mathcal{O}(\nu)
\label{eq:fim_bg}
\end{equation}
However, domain shifts in space situational awareness (e.g., illumination changes) require significant updates $\Delta \theta^*$ along these background-sensitive directions. EWC enforces constraints exactly on this subspace, creating a geometric conflict between adaptation and preservation:
\begin{equation}
\mathcal{L}_{\text{EWC}}(\Delta \theta^*)
\;\approx\;
\frac{1}{2} \sum \lambda_k (\mathbf{v}_k^\top \Delta \theta^*)^2
\;\gg\; 0
\label{eq:ewc_conflict}
\end{equation}
This explains the failure of rigidity-based methods in sparse RSO detection.

\subsection{Accumulated Deviation and Irreplaceability of Dual-Stage Invariance}
\label{app:step4_repair}

Finally, we link the gradient pathology to the prediction error. Let the detector be a composite function $\hat{y} = h_\psi(f_\phi(x))$. 
By invoking the Lipschitz continuity of the head network, the upper bound of the prediction deviation $\|\Delta \hat{y}\|$ under parameter updates decomposes into feature-induced and head-induced terms:
\begin{equation}
\|\Delta \hat{y}\|
\;\le\;
\underbrace{L_h \cdot \|\Delta f_\phi\|}_{\text{Feature Propagation}}
\;+\;
\underbrace{\|\Delta h_\psi\|}_{\text{Head Drift}}
\label{eq:drift_general}
\end{equation}
where $L_h = \|\nabla_f h_\psi\|$ is the Lipschitz constant of the head.

\vspace{2pt}
\noindent 1) Failure of Single-Stage Distillation:
Conventional methods only constrain the head ($\|\Delta h_\psi\| \le \epsilon$). However, as proven in Appendix~\ref{app:nested_sparsity}, the backbone is exposed to background-dominated gradients, leading to unbounded feature drift ($\|\Delta f_\phi\| \gg 0$). Consequently, the error bound in Eq.~\eqref{eq:drift_general} is dominated by the amplified feature noise, termed accumulated deviation:
\begin{equation}
\|\Delta \hat{y}\|_{\text{single}}
\;\approx\;
L_h \cdot \|\Delta f_\phi\|
\;\gg\; \epsilon
\label{eq:single_fail}
\end{equation}

\vspace{2pt}
\noindent 2) Necessity of Dual-Stage Invariance:
Our framework explicitly imposes constraints on both stages ($\|\Delta f_\phi\| \le \epsilon_f, \|\Delta h_\psi\| \le \epsilon_h$). Substituting these into Eq.~\eqref{eq:drift_general} yields a tight stability bound that is independent of background gradient noise:
\begin{equation}
\|\Delta \hat{y}\|_{\text{dual}}
\;\le\;
L_h \cdot \epsilon_f + \epsilon_h
\label{eq:dual_bound}
\end{equation}
This formally proves that dual-stage distillation is a \emph{structural necessity} to suppress accumulated error propagation and prevent catastrophic forgetting under the extreme sparsity regime.

\end{document}